\documentclass[10pt,twocolumn,twoside]{IEEEtran}
\usepackage{cite}
\usepackage[cmex10]{amsmath}
\usepackage{amssymb}
\usepackage{graphicx}
\usepackage{cases}
\usepackage{mdwmath}
\usepackage{mdwtab}
\usepackage{array}
\usepackage{url}
\usepackage{multirow}
\usepackage{booktabs}
\usepackage{threeparttable}
\usepackage{color}
\usepackage{amsmath}

\def\ie{\emph{i.e.}}
\def\eg{\emph{e.g.}}
\def\etal{{\em et al.}}

\newcommand{\mc}[1]{\mathcal{#1}}

\graphicspath{{./figs/}}
\usepackage{color}
\allowdisplaybreaks

\begin{document}
\title{Distortion-adaptive Salient Object Detection in \\ 360$^\circ$ Omnidirectional Images}

\author{Jia Li,~\IEEEmembership{Senior Member,~IEEE,}
        Jinming Su,
        Changqun Xia
        and~Yonghong Tian,~\IEEEmembership{Senior Member,~IEEE}
\thanks{J. Li and J. Su are with the State Key Laboratory of Virtual Reality Technology and Systems, School of Computer Science and Engineering, Beihang University, Beijing, 100191, China. J. Li is also with the Beijing Adavanced Innovation Center for Big Data and Brain Computing, Beihang University, China.}
\thanks{Y. Tian is with the National Engineering Laboratory for Video Technology, School of  Electronics Engineering and Computer Science, Peking University, Beijing, China.}
\thanks{J. Li, C. Xia and Y. Tian are with Peng Cheng Laboratory, Shenzhen, 518000, China.}
\thanks{C. Xia and Y. Tian are corresponding authors. E-mail: xiachq@pcl.ac.cn and yhtian@pku.edu.cn.}
}

\maketitle

\begin{abstract}
Image-based salient object detection (SOD) has been extensively explored in the past decades. However, SOD on 360$^\circ$ omnidirectional images is less studied owing to the lack of datasets with pixel-level annotations. Toward this end, this paper proposes a 360$^\circ$ image-based SOD dataset that contains 500 high-resolution equirectangular  images. We collect the representative equirectangular images from five mainstream 360$^\circ$ video datasets and manually annotate all objects and regions over these images with precise masks with a free-viewpoint way. To the best of our knowledge, it is the first public available dataset for salient object detection on 360$^\circ$ scenes. By observing this dataset, we find that distortion from projection, large-scale complex scene and small salient objects are the most prominent characteristics. Inspired by these foundings, this paper proposes a baseline model for SOD on equirectangular images. In the proposed approach, we construct a distortion-adaptive module to deal with the distortion caused by the equirectangular projection. In addition, a multi-scale contextual integration block is introduced to perceive and distinguish the rich scenes and objects in omnidirectional scenes. The whole network is organized in a progressively manner with deep supervision. Experimental results show the proposed baseline approach outperforms the top-performanced state-of-the-art methods on 360$^\circ$ SOD dataset. Moreover, benchmarking results of the proposed baseline approach and other methods on
360$^\circ$ SOD dataset show the proposed dataset is very challenging, which also validate the usefulness of the proposed dataset and approach to boost the development of SOD on 360$^\circ$ omnidirectional scenes.

\end{abstract}


% \fi
\IEEEpeerreviewmaketitle

\section{Introduction}
\IEEEPARstart{T}{he} purpose of image-based salient object detection (SOD) is to detect and segment objects that capture human visual attention, which is an important preliminary step for various visual tasks such as object recognition~\cite{ren2014region}, tracking~\cite{hong2015online} and image parsing~\cite{lai2016saliency}. In the past decades, many benchmark datasets~\cite{yan2013hierarchical,yang2013saliency,li2014secrets,li2015visual,wang2017learning,xia2017and} have been constructed to drive the development of SOD methods. On these datasets, many learning-based methods~\cite{jiang2013salient,wang2017stagewise,xia2017and,wang2018detect} have been proposed to boot the performance of SOD and have made great progress. These is a fact that almost all existing methods focus on image-/video-based SOD, where the image/video always is displayed with a limited field of view (FoV). However, what human beings perceive is in a three-dimensional world, and the objects analyzed by human vision are 360$^\circ$ omnidirectional scenes at every moment. With the development of imaging technology and hardware, 360$^\circ$ content become more and more widespread on popular image/video sharing platforms. How to analyze 360$^\circ$ content is an emerging problem, which is important to understand and compress image/video information~\cite{ng2005data} as well as enhance user experience. As an important pre-step of most visual tasks, saliency detection is a good tool for 360$^\circ$ image/video analysis.

\begin{figure}[t]
\centering
\includegraphics[width=1.00\columnwidth]{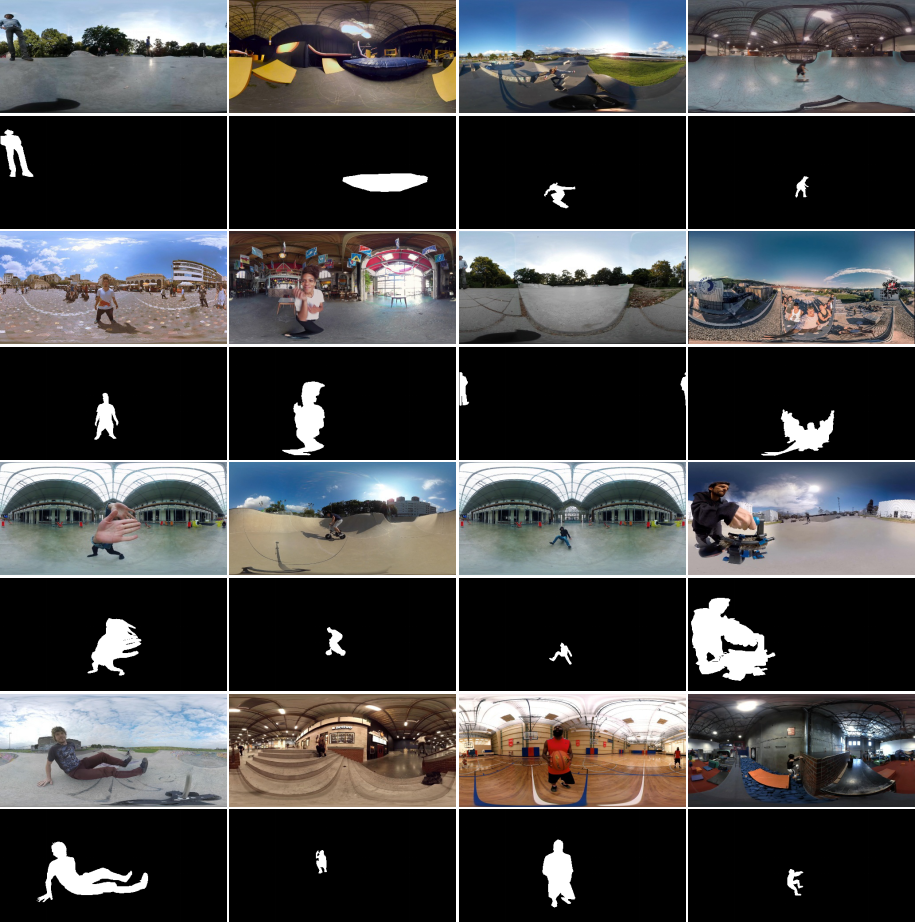}
\caption{Representative examples of \textbf{360-SOD}, shown as equirectangular images and their ground truth.}
\label{fig:360-SOD-examples}
\end{figure}

In recent years, some omnidirectional video datasets~\cite{corbillon2017360,rai2017dataset,xu2017subjective,hu2017deep,cheng2018cube,xu2018gaze,
Zhang_2018_ECCV,xu2018predicting} have appeared. These datasets contain many 360$^\circ$ omnidirectional videos, and are usually used to explore the visual attention mechanism of human beings, mainly human fixation. SOD as a related task of fixation estimation, can better perceive and detect saliency in object-level, and have great significance for higher-level tasks (\eg object tracking). However, there still lack 360$^\circ$ image/video datasets for SOD, which prevents the fast grown of this branch. Judging from the actual needs, in order to meet the needs of analyzing the growing omnidirectional imaging data by means of computer technology, it is necessary to construct a 360$^\circ$ datasets for SOD to promote the development of SOD on omnidirectional data. 

Toward this end, this paper proposes \textbf{360-SOD}, a 360$^\circ$ omnidirectional image dataset for SOD, which contains 500 images with pixel-level annotation including various scenes as some representative examples shown in Fig.~\ref{fig:360-SOD-examples}. In constructing \textbf{360-SOD}, we first collect a large number of 360$^\circ$ videos, and segment them to get 6870 key frames. Given these images, we ask two volunteers to judge whether there contain unambiguous salient objects and whether this frame is a redundancy of adjacent image, and finally 500 images with obvious salient objects are picked out. In the next stage, six engineers are asked to manually label the accurate boundaries of salient objects. The judgment of salient objects is done in omnidirectional picture brower, and the annotation is based on the equirectangular form of 360$\circ$ omnidirectional images. To the best of our knowledge, \textbf{360-SOD} is the first dataset for salient object detection on 360$^\circ$ scenes.

\begin{figure}[t]
\centering
\includegraphics[width=1.00\columnwidth,height=7.5cm]{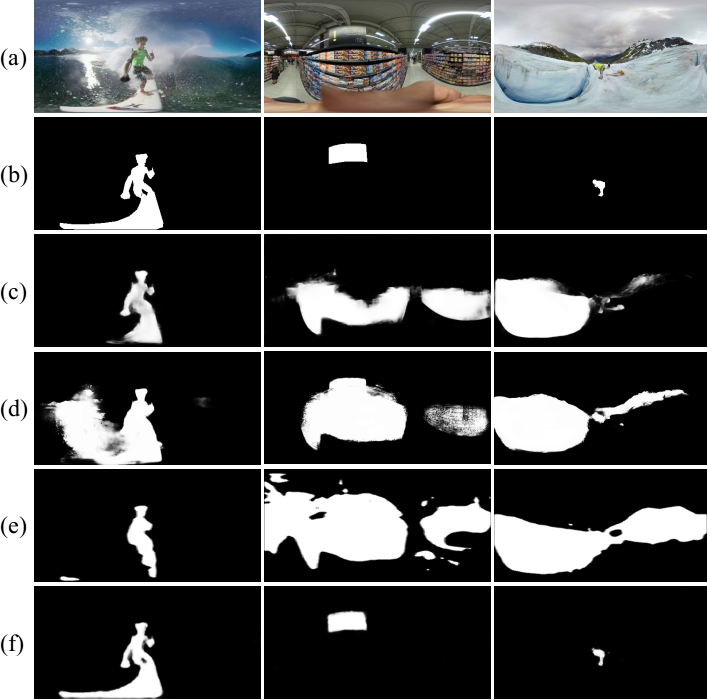}
\caption{Examples of predicted salient objects by different methods. (a) equirectangular form of 360$^\circ$ images, (b)ground truth, and saliency maps of (c) RAS~\cite{chen2018reverse}, (d) R3Net~\cite{deng2018r3net}, (e) DGRL~\cite{wang2018detect} and (f) the proposed approach.}
\label{fig:DDS-motivation}
\end{figure}

Based on the omnidirectional image dataset, we explore SOD on 360$^\circ$ images. To predict the salient objects on 360$^\circ$ imaging data, an intuitive method is to store the 360$^\circ$ scenes as equirectangular images, and then directly utilize the SOD algorithms on conventional images. However, the predicted saliency map is not satisfactory as shown in Fig.~\ref{fig:DDS-motivation}. By exploring the difference between conventional images and 360$^\circ$ omnidirectional images, we find there are mainly three problems in the processing of omnidirectional images. The first problem is the distortion caused by projections from sphere to plane as shown in the first column of Fig.~\ref{fig:DDS-motivation}. No matter which projection method (\eg equirectangular, cube map and patch-based projections) we choose, distortion is inevitable. In this work, we choose equirectangular projection to store and analyze these images as done in ~\cite{su2016pano2vid,hu2017deep}. The second problem is caused by large-scale complex scene in omnidirectional image as depicted in the second column of Fig.~\ref{fig:DDS-motivation}, which may confuse algorithms to detect wrongly. The last problem is the difficulty of perceiving and segmenting small salient objects as presented in the third column of Fig.~\ref{fig:DDS-motivation}. These problems make it difficult to deal with SOD on omnidirectional images.

To address these issues, we propose a baseline model on \textbf{360-SOD} to consider the basic problems in omnidirectional scenes. In the proposed approach, we construct a distortion-adaptive module to deal with the distortion due to equirectangular projection. And then, the approach introduces a multi-scale contextual integration module to undertake the large-scale complex scene. Moreover, this model is organized in a progressive refinement manner to restore the boundaries and small salient objects. Experimental results show the proposed method baseline model (denoted as \textbf{DDS}) outperforms the three top-performanced conventional SOD methods on \textbf{360-SOD}.

The contributions are summarized as follows: 1) we propose the first SOD dataset on 360$^\circ$ omnidirectional images, which we will release to boost the development of 360$^\circ$ SOD, 2) we propose a baseline model to deal with the basic problems in omnidirectional scenes and it outperforms top-performanced conventional methods on 360$^\circ$ SOD dataset, 3) we provide a comprehensive benchmark of our approach and other state-of-the-art SOD methods on 360$^\circ$ scenes, which reveals the key challenges in omnidirectional scenes and validates the usefulness of the proposed dataset and baseline model.

The rest of this paper is organized as follows: Section \uppercase\expandafter{\romannumeral2} reviews existing datasets and models about SOD. Section \uppercase\expandafter{\romannumeral3} presents the new dataset on 360$^\circ$ omnidirectional images. In Section \uppercase\expandafter{\romannumeral4}, a baseline model for SOD on 360$^\circ$ images is proposed. Section \uppercase\expandafter{\romannumeral5} benchmarks the proposed model and other state-of-the-art methods, and the paper is concluded in Section  \uppercase\expandafter{\romannumeral6}.

\section{Related work}

\begin{table*}[t]
\centering
\caption{Comparison between representative 360$^\circ$ image/video datasets. \#Image/Video: the number of image/video, \#Frames: the total number of frames and \#Time: the total number of seconds.}
\setlength{\tabcolsep}{2mm}{
\renewcommand\arraystretch{1.2}
\begin{tabular}{c|c|c c|c c c|c c}
\hline
\multirow{2}*{Dataset} & \multirow{2}*{Type of Scene} & \multirow{2}*{\#Image} & \multirow{2}*{\#Video} & \multicolumn{3}{c|}{\#Resolution(in pixels)} & \multirow{2}*{\#Frames} & \multirow{2}*{\#Time (in seconds)}   \\
& & & & Width & Height & Max Resolution &    \\
\hline
360-VHMD~\cite{corbillon2017360}
            & indoor \& outdoor & - & 7 & [3840, 3840] & [2048, 2160] & 3840 $\times$ 2160       & 48,414 & 1,340\\
            
Salient!360~\cite{rai2017dataset} & indoor \& outdoor & 85 & 19 & [3840, 18332] & [1920, 9166] & 18332 $\times$ 9166 & 10,548 & 381\\

Wild-360~\cite{cheng2018cube} &outdoor& - &  85 & [1920, 2160] & [960, 1080] & 2160 $\times$ 1080 & 40,290 & 1,553\\

VR-scene~\cite{xu2018gaze} & indoor \& outdoor & - & 208 & [1920, 7680] & [1080, 3840] & 7680 $\times$ 3840 & 215,457 & 7,511\\

360-saliency~\cite{Zhang_2018_ECCV} & indoor \& outdoor & - &  104 & [3724, 3840] & [1862, 2160] & 3840 $\times$ 2160 & 76,611 & - \\

VR-VQA48~\cite{xu2017subjective} & indoor \& outdoor & - & 48 & [2880, 7680]  & [1440, 3840]  & 7680 $\times$ 3840 & 35,906  & 1326 \\

Sports-360~\cite{hu2017deep} & indoor \& outdoor & - & 342 & [3724, 3840] & [1862, 2160] & 3840 $\times$ 2160 & 180,000 & -  \\

PVS-HM~\cite{xu2018predicting} & indoor \& outdoor & - & 76 & [2880, 7680]  & [1440, 3840]  & 7680 $\times$ 3840 & - & 2,045 \\
\hline
\end{tabular}}
%\begin{tablenotes}
%       			\footnotesize
%       			\item[1] $^*$ indicates the value of resolution, the number of frames or time are estimates. 
%       		\end{tablenotes}
\label{tab:dataset-statistic}
\end{table*}

SOD on 360$^\circ$ omnidirectional images is correlated with image-/video-based SOD, and 360$^\circ$ datasets and tasks. In this section, we will review the most related datasets and models.

\subsection{Conventional Datasets for SOD}
There are many image datasets.
ECSSD~\cite{yan2013hierarchical} contains 1,000 images with complex structures and obvious semantically meaningful objects. 
DUT-OMRON~\cite{yang2013saliency} consists of 5,168 complex images with pixel-wise annotations and all images are downsampled to a maximal side length of 400 pixels. 
PASCAL-S~\cite{li2014secrets} comprises 850 natural images that are pre-segmented into objects or regions and free-viewed by 8 subjects in eye-tracking tests for salient object annotations. 
HKU-IS~\cite{li2015visual} includes 4,447 images and lots of images contain multiple disconnected salient objects or salient objects that touch image boundaries. 
DUTS~\cite{wang2017learning} is a large scale dataset containing 10,533 training images and 5,019 testing images. The images are challenging with salient objects that occupy various locations and scales as well as complex background. 
XPIE~\cite{xia2017and} has 10,000 images covering a variety of simple and complex scenes with salient objects of different numbers, sizes and positions. These datasets with pixel-level annotation drive the development of learning-base model~\cite{jiang2013salient,wang2017stagewise,xia2017and,wang2018detect} for image-based SOD.

Many video datasets have been proposed for SOD.
SegTrack V2~\cite{li2013video} is a classic dataset in video object segmentation that is frequently used in many previous works. It consists of 14 densely annotated video clips with 1,066 frames in total. 
Youtube-Objects~\cite{prest2012learning} contains a large number of Internet videos and its widely used subset~\cite{jain2014supervoxel} contains 127 videos with 20,977 frames. In these videos, 2,153 key frames are sparsely sampled and manually annotated with pixel-wise masks according to the video tags. 
VOS~\cite{li2018benchmark} contains 200 videos with 116, 093 frames. On 7,467 uniformly sampled key frames, all objects are pre-segmented by 4 subjects, and the fixations of another 23 subjects are collected in eye-tracking tests. These datasets provide sufficient data support for video-based SOD algorithms~\cite{wang2015consistent,li2017primary}.

\subsection{Conventional Models for SOD}
Hundreds of image-based SOD methods have been proposed in the past decades. The survey~\cite{borji2015salient} provides detailed introduction and analysis about SOD methods, especially traditional methods. Recently, a lot of deep models are devoted to enhance the performance of neural networks for SOD.
Zhang~\etal~\cite{zhang2018progressive} proposed an attention guided network to selectively integrates multi-level information in a progressive manner. 
Wang~\etal~\cite{wang2017stagewise} proposed a pyramid pooling module and a multi-stage refinement mechanism to gather contextual information and stage-wise results, respectively
Chen~\etal~\cite{chen2018reverse} proposed reverse attention mechanism which is inspired from human perception process by using top information to guide bottom-up feed-forward process in a top-down manner. 
Chen~\etal~\cite{chen2017look} incorporated human fixation with semantic information to simulate the human annotation process for salient objects.

With the development of image-based SOD, video-based SOD also has made great progress. 
Papazoglou and Ferrari~\cite{papazoglou2013fast} proposed an approach for the fast segmentation of foreground objects from background regions. They estimated an initial foreground map with respect to the motion information, which was then refined by building the foreground/background appearance models and encouraging the spatiotemporal smoothness of foreground objects across the whole video. 
Wang~\etal~\cite{wang2015saliency} proposed an unsupervised algorithm for video-based SOD. In their algorithm, frame-wise saliency maps were first generated and refined with respect to the geodesic distances between regions in the current frame and subsequent frames. After that, global appearance models and dynamic location models were constructed so that the spatially and temporally coherent salient objects can be segmented. 
Li~\etal~\cite{li2017primary} proposes an approach for segmenting primary video objects by using Complementary Convolutional Neural Networks (CCNN) and neighborhood reversible flow. In their approach, the initialized foregroundness and backgroundness can be efficiently and accurately propagated along the temporal axis so that primary video objects gradually pop-out and distractors are well suppressed.

\subsection{Panoramic Datasets}
360-VHMD~\cite{corbillon2017360} is a popular dataset for 360$^\circ$ videos. It contains 7 videos about indoor and outdoor scenes with 48414 frames. 
Salient!360~\cite{rai2017dataset} contains 85 equirectangular images and 19 equirectangular videos with ground-truth fixation maps and scan-paths obtained from subjective experiments. 
Wild-360~\cite{cheng2018cube} consists of 85 360$^\circ$ video clips, totally about 40k frames. 60 clips within wild-360 are for training and the rest 25 clips are for testing. All the clips are cleaned and trimmed from 45 raw videos obtained from YouTube. 
VR-scene~\cite{xu2018gaze} has 208 high definition dynamic 360$^\circ$ videos collected from Youtube, each with at least 4k resolution (3840 pixels in width) and 25 frames per second. The duration of each video ranges from 20 to 60 seconds. The videos in VR-scene exhibit a large diversity in terms of contents, which include indoor scene, outdoor ac-
tivities, music shows, etc. Further, some videos are captured from a fixed camera view and some are shot with a moving camera. 
360-saliency~\cite{Zhang_2018_ECCV} collects 104 video clips as the data used for saliency detection in 360$^\circ$ videos. The video contents involve five sports (i.e. basketball, parkour, BMX, skateboarding, and dance), and the duration of each video is between 20 and 60 seconds. 

In addition, VR-VQA48~\cite{xu2017subjective} is a 360$^\circ$ dataset for measuring the quality reduction of panoramic videos. It contains viewing direction data of 40 subjects on 48 sequences of panoramic videos. All of these sequences are downloaded from YouTube and VRCun and the resolution is beyond 3K and up to 8K. 
Sports-360~\cite{hu2017deep} consists of  342 360$^\circ$ videos downloaded from YouTube in five sports domains including basketball, parkour, BMX,
skateboarding, and dance, which is created for the study on relieving the viewer from this ``360 piloting'' task. 
PVS-HM~\cite{xu2018predicting} is a new panoramic video database that consists of head movement (HM) positions across 76 panoramic video sequences with a thorough analysis. These panoramic video sequences are from YouTube and VRCun, and the duration of each sequence is cut to be from 10 to 80 seconds.

%It is likely that the quality of experience of panoramic videos dramatically degrades due to compression artifacts or low resolutions. In order to measure the quality reduction of panoramic videos, Xu~\etal~\cite{xu2017subjective} establish a viewing direction database for panoramic videos and propose a new method for subjective visual quality assessment (VQA) on panoramic videos.
 
%For another common scenario, watching a 360$^\circ$ sports video requires a viewer to continuously select a viewing angle. To relieve the viewer from this ``360 piloting'' task, Hu~\etal~\cite{hu2017deep} propose a deep learning-based agent for piloting through 360$^\circ$ sports videos automatically and they build first 360$^\circ$ sports videos dataset to train and evaluate the proposed “deep 360 pilot” agent.

%In addition, head movement (HM) plays a key role in modeling human attention on panoramic video. Xu~\etal~\cite{xu2018predicting} establish a new panoramic video database that consists of HM positions across 76 panoramic video sequences with a thorough analysis and they propose a deep reinforcement learning-based HM prediction approach with offline and online versions.

A comprehensive statistic of these datasets is listed in Tab.~\ref{tab:dataset-statistic}. These datasets almost contain annotation of human fixation, while there is no annotation of salient objects. Although a large number of approaches for SOD have been proposed, there is no approach to focus on SOD on omnidirectional images/videos. In addition, even if many omnidirectional datasets have been constructed, there are few 360$^\circ$ datasets for the task of SOD on 360$^\circ$ scenes.

\section{A New Dataset for 360$^\circ$ SOD}
In this section, we will introduce the rules and details in constructing the 360$^\circ$ image dataset.

\begin{figure}[t]
\centering
\includegraphics[width=1.00\columnwidth]{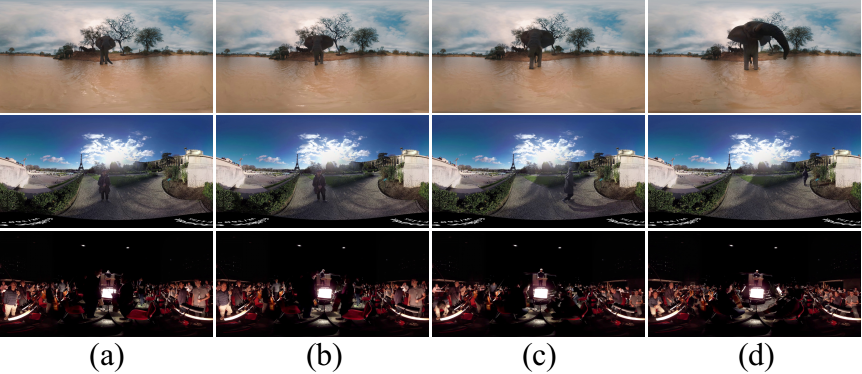}
\caption{Examples of adjacent four key frames in raw 360$^\circ$ image dataset.}
\label{fig:raw-images}
\end{figure}

\begin{figure}[t]
\centering
\includegraphics[width=1.00\columnwidth]{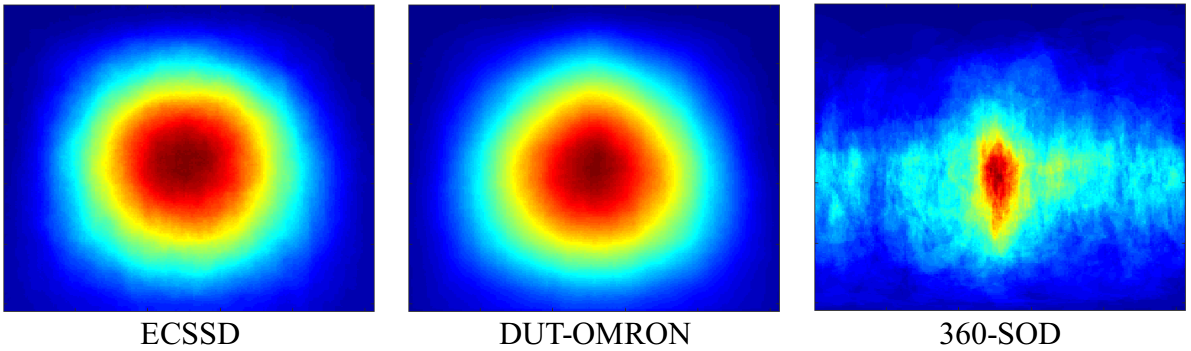}
\caption{Average annotation maps of two convolutional image-based SOD datasets and \textbf{360-SOD}.}
\label{fig:jetmap}
\end{figure}

\subsection{Data Collection}
As introduced in Section \uppercase\expandafter{\romannumeral2}, there exist many omnidirectional video datasets. 360-VHMD~\cite{corbillon2017360}, Salient!360~\cite{rai2017dataset}, Wild-360~\cite{cheng2018cube}, VR-scene~\cite{xu2018gaze} and 360-saliency~\cite{Zhang_2018_ECCV} are almost for the research of visual attention and contain the ground truth of human fixation. Therefore, there datasets are likely to have regions or objects that can attract human visual attention. Instead of undertaking a new data shooting and video recording, we directly combine these five datasets as a raw data source. Next, we intercept a key frame every two seconds from these videos in the data source. We collect these key frames and merge them with the original images. After that, we resize each image to have a maximum side length of 1024 pixels for convenient processing. Finally, we obtain a raw 360$^\circ$ omnidirectional image dataset with 6870 images. Note that because the 360-saliency doesn't provide the video frame rate, we adopt 30FPS as its default value.

\subsection{Annotation of Salient Objects}
There exist some scenes with unclear salient objects and redundant data between continuous key frames in above raw 360$^\circ$ image dataset, as shown in Fig.~\ref{fig:raw-images}. Therefore, we divide the process of annotation of this dataset into two stages. In the first stage, these 6870 images are displayed in omnidirectional picture brower in original order. Then, we ask two volunteers to judge whether an image is selected according to three main principles, which include: (1) there is clear scene without jitter, (2) unambiguous, meaningful and annotatable salient objects exist in the scene, and (3) if the adjacent multiple images has same salient objects with similar shape and position, only the best one is selected. If the answer is ``Yes'', we will collect this image. After the first stage that 6870 360$^\circ$ images are processed, we finally collect 500 images from the raw 360$^\circ$ image dataset. In the second stage, six engineers are asked to judge the salient objects in omnidirectional picture brower and manually label the accurate boundaries of salient objects by LabelMe Annotation Tool~\cite{russell2008labelme} in the equirectangular form of these 500 images. The manual annotation is a time-consuming work, and it takes an average of five minutes to annotate each image. Note that we have two volunteers involved in the process for cross-check the quality of annotations. 

\begin{figure}[t]
\centering
\includegraphics[width=1.00\columnwidth]{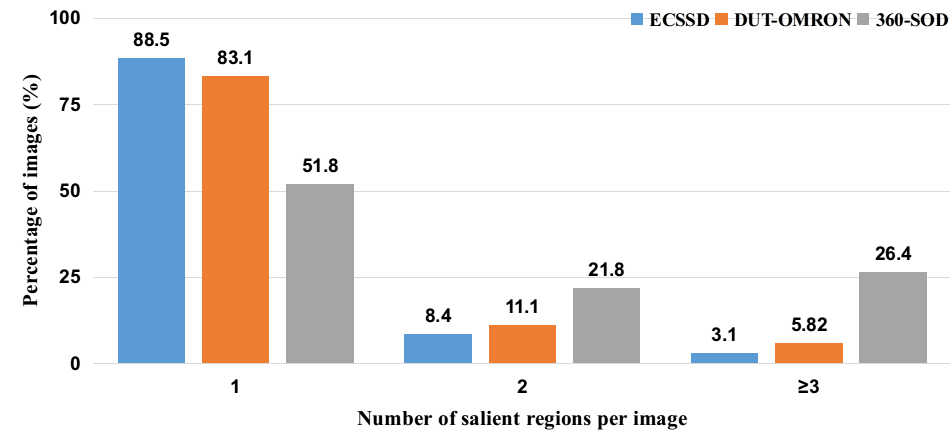}
\caption{Histograms of the number of salient objects in 3 datasets.}
\label{fig:statistic1}
\end{figure}

\begin{figure}[t]
\centering
\includegraphics[width=1.00\columnwidth]{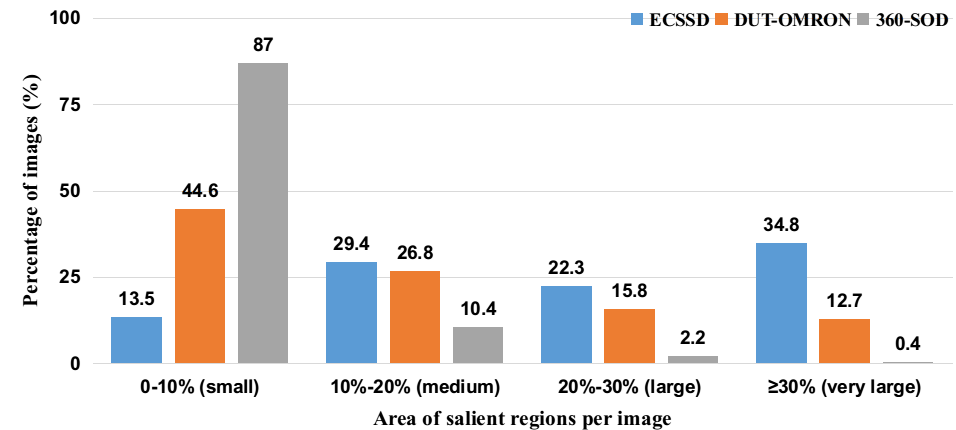}
\caption{Histograms of the area of salient objects in 3 datasets.}
\label{fig:statistic2}
\end{figure}

\begin{figure*}[t]
\centering
\includegraphics[width=1.00\textwidth]{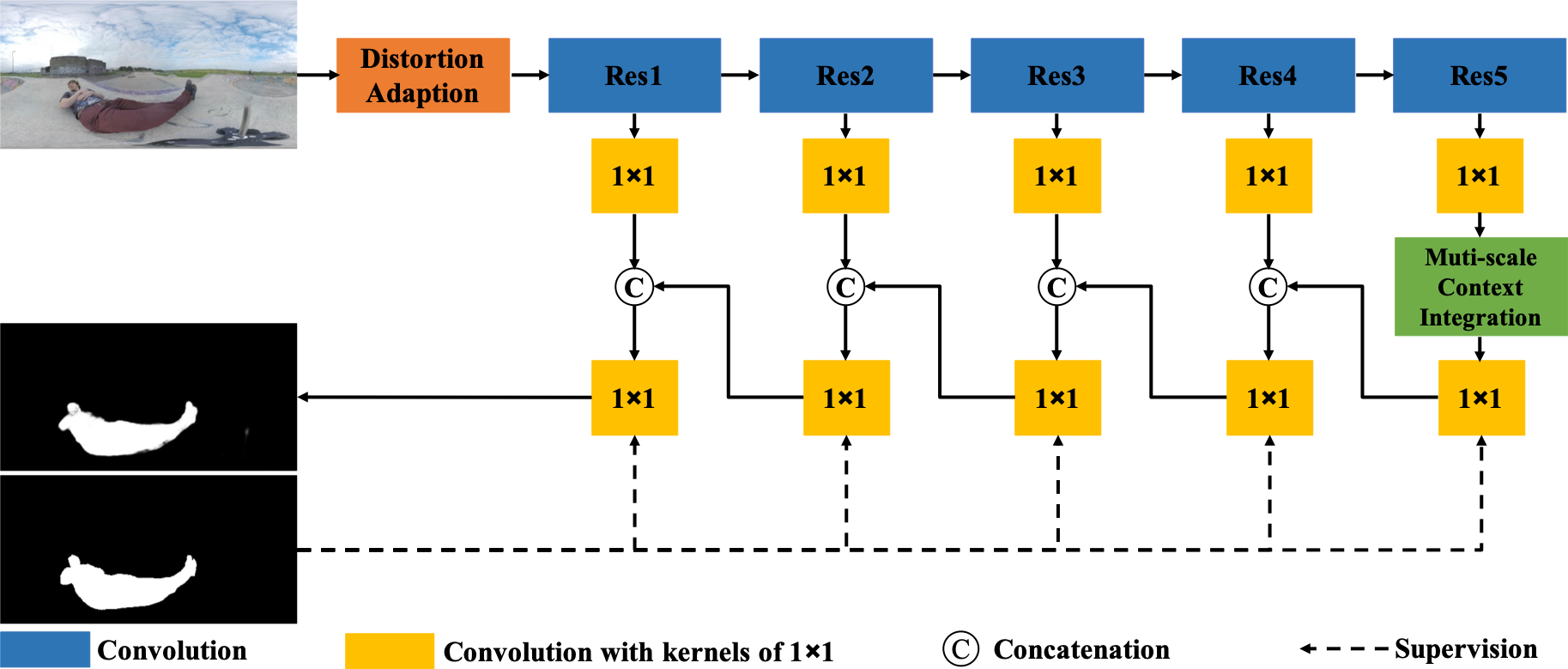}
\caption{The framework of our baseline model. The equirectangular image is processed by a distortion-adaptive module, and the output is transferred to ResNet-50 to extract features in different levels. In the highest-level features are dealt with a multi-scale context integration module to integrate multi-scale contextual information. Moreover, the coarser-level saliency feature is concatenated into the adjacent finer-level features to get finer saliency map and the whole network is organized in a progressive form.}
\label{fig:framework}
\end{figure*}

Finally, we obtain the final 360$^\circ$ image dataset, which contains 500 equirectangular images with pixel-level annotation. This dataset includes indoor and outdoor scene from different perspectives as well as different scene complexities. The dataset is denoted as \textbf{360-SOD}. Some representative examples of \textbf{360-SOD} can be found in Fig.~\ref{fig:360-SOD-examples}.

\subsection{Dataset Statistics}
To explore the main characteristics of \textbf{360-SOD}, we present the annotation maps (AAMs) of \textbf{360-SOD} and two conventional image-based SOD datasets (\ie, ECSSD~\cite{yan2013hierarchical} and DUT-OMRON~\cite{yang2013saliency}) as shown in Fig.~\ref{fig:jetmap}. As in~\cite{borji2015salient}, the AAM of an image-based SOD dataset is computed by 1) resizing all ground-truth masks from the dataset to the same resolution, 2) summing the resized masks pixel by pixel, and 3) normalizing the resulting map to a maximum value of 1.0. In this way, the figure gives a better view of the distribution of salient objects in all the images in \textbf{360-SOD}.

From Fig.~\ref{fig:jetmap}, we can see that the distributions of convolutional image-based SOD datasets are usually center-biased, while the distribution of the 360$^\circ$ omnidirectional image-based dataset \textbf{360-SOD} is more discrete. This may be caused by the characteristic of large-scale complex scenes in 360$^\circ$ image, which indicates the diversity of \textbf{360-SOD} and the difficulty of 360$^\circ$ SOD.

In addition, we present the histograms of number and area of salient objects in \textbf{360-SOD} and other two datasets in Fig.~\ref{fig:statistic1} and Fig.~\ref{fig:statistic2}. As shown in Fig.~\ref{fig:statistic1}, we can find that there are usually more salient objects in \textbf{360-SOD} that other two datasets, which is caused by the difference of conventional and 360$^\circ$ images. Moreover, we can see there are more images with small areas of salient objects in Fig.~\ref{fig:statistic2}, which maybe one of the main challenges in 360$^\circ$ omnidirectional images.

\subsection{Analysis}
We directly deal with this dataset by existing conventional SOD algorithm and the results are presented in Fig.~\ref{fig:DDS-motivation}, which are obviously not satisfactory. We explore the difference between conventional images and 360$^\circ$ omnidirectional images, and we find there are three main problems leading to the difficulty in 360$^\circ$ SOD: 1) distortion, which is inevitable because of the projection from sphere to plane; 2) large-scale complex scene, which is the main characteristics of omnidirectional images; and 3)small salient objects that commonly exist in 360$^\circ$ images as presented in Fig.~\ref{fig:DDS-motivation}. These issues actually prevent the detection of salient objects in 360$^\circ$ images.

\section{A baseline model of SOD on 360$^\circ$ images}

To deal with the three existing issues in omnidirectional images (\ie distortion from projection, large-scale complex scene and small salient objects), we propose a baseline model for SOD on 360$^\circ$ images. The baseline model is inspired by the existing issues, and it is fed by the equirectangular image as input and outputs a saliency map with the same resolution as input as presented in Fig.~\ref{fig:framework}. Details of the proposed approach are descried as follows.

\subsection{Architecture}
As depicted in Fig.~\ref{fig:360-SOD-examples}, the baseline model is a distortion-adaptive network with deep supervision (denoted as \textbf{DDS}). The first module of \textbf{DDS} is a distortion-adaptive module, which is designed to deal with the distortion caused by the projections from sphere to plane. After the processing of distortion-adaptive module, the distortion in the image will be corrected adaptively and a new image is output. Following this module, \textbf{DDS} takes ResNet-50~\cite{he2016deep} as the feature extractor, which is modified to remove the last global pooling and fully connected layers for the pixel-level prediction. Feature extractor has five residual modules, named as $\mc{R}_{1}, \mc{R}_{2}, ..., \mc{R}_{5}$. To obtain larger feature maps, the strides of all convolutional layers belonging to last two residual modules $\mc{R}_{4}$ and $\mc{R}_{5}$ are set to 1. To further enlarge the receptive fields of high-level features, we set the dilation rates~\cite{yu2015multi} to 2 and 4 for convolutional layers in $\mc{R}_{4}$ and $\mc{R}_{5}$, respectively. For a $H \times W$ input images, a $\frac{H}{8} \times \frac{W}{8}$ feature map is output by the feature extractor.

\begin{figure}[t]
\centering
\includegraphics[width=0.80\columnwidth]{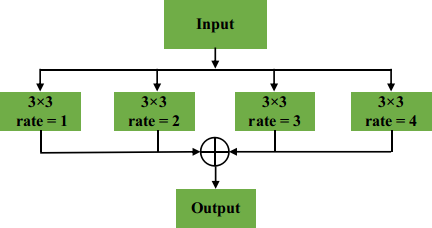}
\caption{Structures of the multi-scale context integration module.}
\label{fig:MCI}
\end{figure}

We select the outputs of last convolutional layers in $\mc{R}_{1}, \mc{R}_{2}, ..., \mc{R}_{5}$ as side-outputs as used in ~\cite{hou2019deeply,chen2018reverse}. For each side-output, its features are processed by a convolutional layer with kernels $1 \times 1$ to compress the feature channel. For the side-output of $\mc{R}_{5}$, after the channel compression layer, a multi-scale context integration module is conducted to integrate multi-scale contextual information, which is for undertaking the large-scale complex scene. The multi-scale context integration module is a variant of Atrous Spatial Pyramid Pooling (ASPP)~\cite{chen2018deeplab} as shown in Fig.~\ref{fig:MCI}. The module is composed of four branches with four dilation convolutional layers. The four layers consist of 128 kernels of $3 \times 3$ with dilation rates of 1, 2, 3 and 4, followed by an element-wise summation operation. The multi-scale context integration module is able to effectively integrate multi-scale features. Following this module, another convolutional layer with one kernel $1 \times 1$ to convert the feature space from high dimension to saliency feature with one channel. Moreover, the saliency feature is upsampled and concatenated into the output of $\mc{R}_{4}$ to get finer saliency feature. This mechanism of coarser-level saliency feature is concatenated into the adjacent finer-level features are conducted on every two adjacent side-outputs, and the whole network is organized in a progressive form.
 
\subsection{Distortion Adaptation for Equirectangular Images}
\begin{figure}[t]
\centering
\includegraphics[width=1.00\columnwidth]{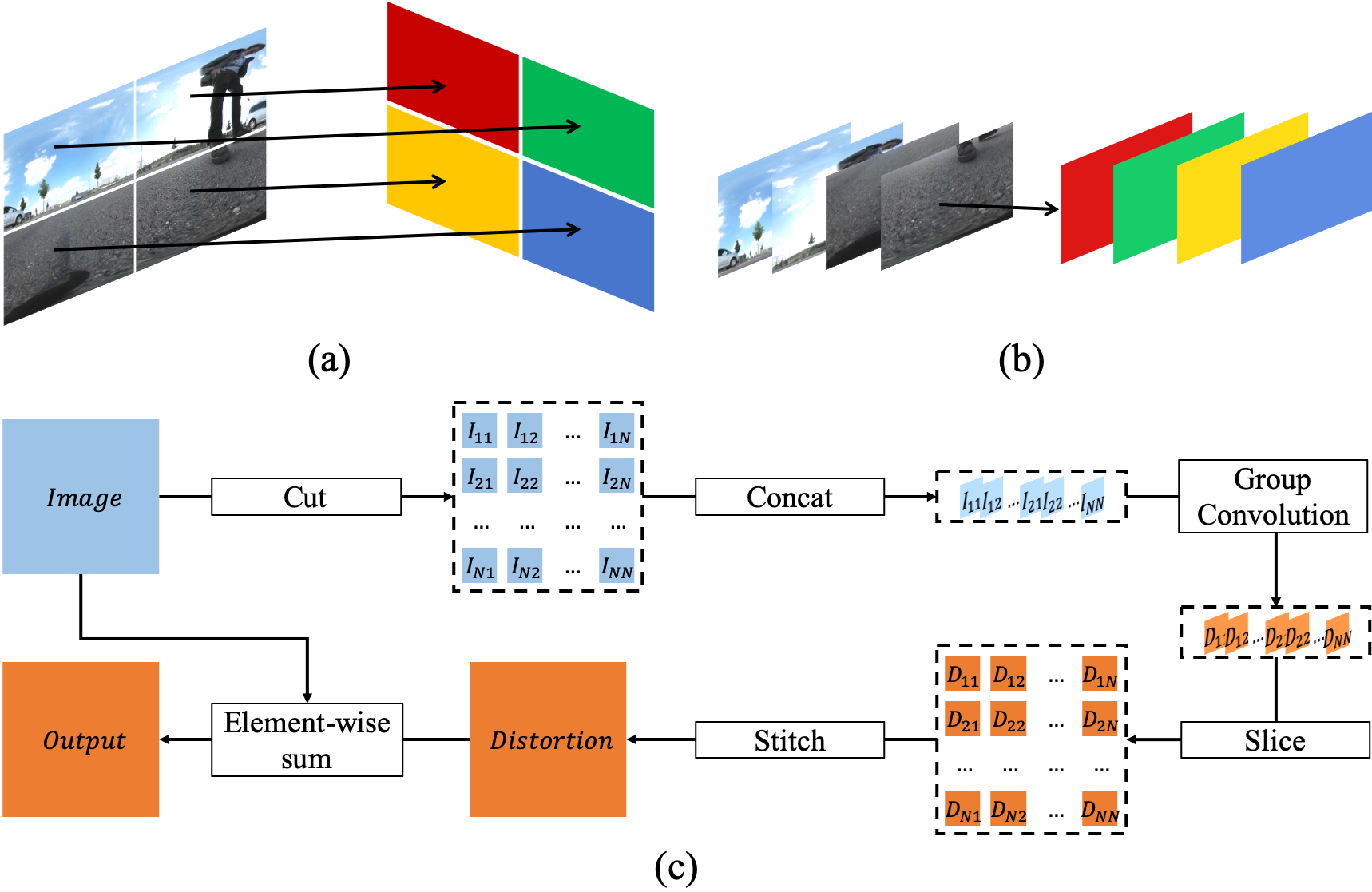}
\caption{Details of distortion-adaptive module. In this example, the image is cut into 4 image blocks, and the 4 blocks are convolved with 4 different convolutional kernels that are display as 4 colors in this figure. The convolutional kernels are learnable. Note we choose the number of image blocks as 4 for convenient display. (a) and (b) are equivalent forms, and (c) is the complete structure of distortion-adaptive module, where $I_{ij}$, $D_{ij}$, $N$ and other details are introduced in Section VI-B.}
\label{fig:equivalent-form-of-DA}
\end{figure}
To deal with the projection distortion in equirectangular images, a distortion-adaptive module is conducted to correct the distortion of the input images adaptively. The equirectangular projection is one of most commonly used projection methods from sphere to plane, which will bring different degrees of distortion in different positions of the image, especially locations near the poles.

To address this distortion, we propose a distortion-adaptive module to deal with different regions with various parameters. As shown in Fig.~\ref{fig:equivalent-form-of-DA}(a), we cut the input equirectangular image into $N \times N$ image blocks and we use $I_{ij} \ (i,j = 1,2, \cdots, N)$ that represents the image block at $i$th row and $j$th column, and $K_{ij}$ which represents the corresponding kernels of $I_{ij}$. In this manner, $I_{ij}$ is only convolved with the $K_{ij}$ and $K_{ij}$ is only used to convolve $I_{ij}$. Moreover, $K_{ij}$ has three channels and $I_{ij}$ has three kernels with three channels. In this module, different image blocks are convolved with different kernels. 
The output has the same resolution as input, and it also can be regarded as $N \times N$ image blocks, which is the learned distortion. We denote the distortion at $i$th row and $j$th column of output as $D_{ij}$, so $D_{ij}$ is computed by 
\begin{equation}
D_{ij} = I_{ij} \ast K_{ij},
\end{equation}
where $\ast$ mean the standard convolution operation. 

In a specific implementation, we can conduct this operation by group convolution operation as presented in Fig.~\ref{fig:equivalent-form-of-DA}(b). In detail, we first cut the input image into $N \times N$ image blocks with 3 channels, and then we concatenate these image blocks in the dimension of channel. Next, a group convolutional layer with $N \times N$ kernels is conducted, which is equivalent to directly convolve different image blocks with different kernels. The last step is to slice the learned distortion as the reverse operation of concatenation as well as stitch image blocks in the reverse manner of cutting operation. In particular, we organize the distortion-adaptive module in residual learning whose complete structure is depicted in Fig.~\ref{fig:equivalent-form-of-DA}(c). $O_{ij}$ in output can be computed by
\begin{equation}
O_{ij} = I_{ij} + I_{ij} \ast K_{ij}.
\end{equation}

Then, this output is further encoded and decoded by the following neural network, and finally supervised by the ground truth of salient objects. In this process, all parameters in the whole network (including parameters in the convolutional layer of the distortion-adaptive module) are constrained by the supervisory signal of the ground truth of salient objects, so as to realize parameter learning through gradient back-propagation. Therefore, through end-to-end training of the model, the distortion-adaptive module can be supervised and learned.

%Distortion-adaptative module is learnable and contains learnable parameters from the group convolution layer. This module is indirectly supervised by the ground truth of salient objects. In this process, distortion in different locations of the image will be adaptively learned and corrected.

\begin{table*}[t]
\centering
\caption{Performance benchmark of 13 state-of-the-art models before being fine-tuned on \textbf{360-SOD}. The best three results are in {\color{red}{\textbf{\underline{red}}}}, {\color{green}{\textbf{\underline{green}}}} and {\color{blue}{\textbf{\underline{blue}}}}. }
\setlength{\tabcolsep}{0.38mm}{
\renewcommand\arraystretch{1.2}
\begin{tabular}{c | c c c c c c c c c c c c c}
\hline
 & ELD\cite{lee2016deep}
 & UCF\cite{zhang2017learning}
 & NLDF\cite{luo2017non}
 & Amulet\cite{zhang2017amulet}
 & FSN\cite{chen2017look}
 & SRM\cite{wang2017stagewise}
 & C2SNet\cite{li2018contour}
 & RAS\cite{chen2018reverse}
 & PiCANet\cite{liu2018picanet}
 & R3Net\cite{deng2018r3net}
 & DGRL\cite{wang2018detect}
 & RFCN\cite{wang2018salient}
 & DSS\cite{hou2019deeply}\\
\hline
MAE $\downarrow$
& 0.135 & 0.237 & {\color{green}{\textbf{\underline{0.089}}}} & 0.191 & 0.115 & 0.123 & 0.144 & {\color{red}{\textbf{\underline{0.079}}}} & 0.133 & 0.101 & 0.135 & 0.103 & {\color{blue}{\textbf{\underline{0.094}}}} \\
F$^w_\beta$ $\uparrow$
& 0.213 & 0.203 & 0.339 & 0.226 & 0.289 & 0.302 & 0.266 & {\color{red}{\textbf{\underline{0.395}}}} & 0.327 & {\color{blue}{\textbf{\underline{0.341}}}} & {\color{green}{\textbf{\underline{0.342}}}} & 0.270 & 0.338 \\
F$_\beta$ $\uparrow$
& 0.234 & 0.248 & 0.369 & 0.260 & 0.321 & 0.357 & 0.290 & {\color{red}{\textbf{\underline{0.417}}}} & 0.363 & {\color{blue}{\textbf{\underline{0.408}}}} & {\color{green}{\textbf{\underline{0.415}}}} & 0.325  & 0.356\\ 
\hline
\end{tabular}}
\label{tab:performance_before}
\end{table*}

\begin{figure*}[t]
\centering
\includegraphics[width=1.00\textwidth]{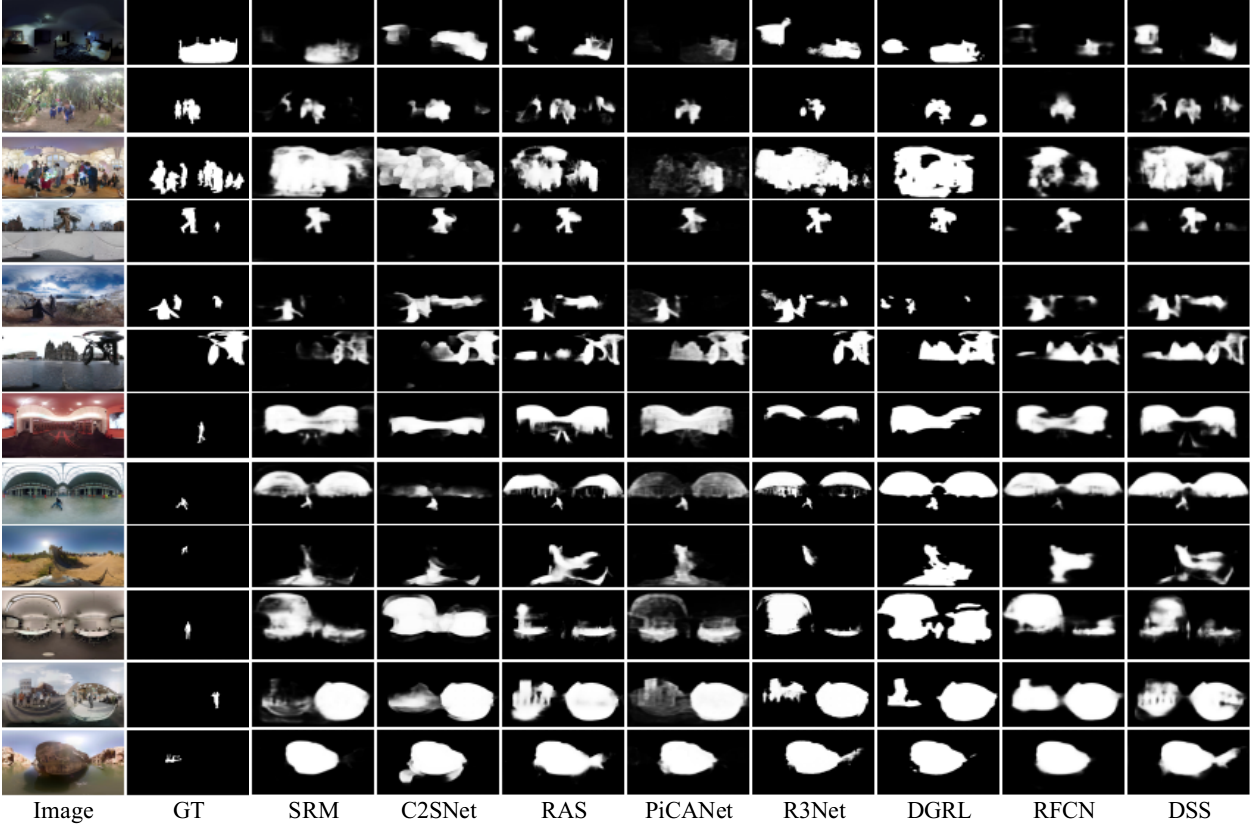}
\caption{Representative examples of the state-of-the-art algorithms on \textbf{360-SOD} without fine-tuning.}
\label{fig:results-before}
\end{figure*}

\subsection{Supervision}
We adopt deep supervision applied to each side-output as used in ~\cite{chen2018reverse,hou2019deeply}. In our network, each side-output is penalized by a standard binary cross-entropy loss. Formally, we denote the cross-entropy loss function of $s$th side-output as $\mathcal{L}^{(s)}$, which is computed by the following formulation: 

\begin{equation}
\begin{split}
& \mathcal{L}^{(s)}(\mathbf{I}, \mathbf{G}, \mathbf{W}, \mathbf{w}^{(s)}) \\
& = - \sum_{l=1}^{|\mathbf{I}|} \mathbf{G}(l)\log(P(\mathbf{G}(l) = 1| \mathbf{I}(l); \mathbf{W}, \mathbf{w}^{(s)}))  \\
& \ \quad +  (1 - \mathbf{G}(l))\log(P(\mathbf{G}(l) = 0| \mathbf{I}(l); \mathbf{W}, \mathbf{w}^{(s)})),
\end{split}
\end{equation}
where $\mathbf{I}$ and $\mathbf{G}$ represent the input image and the corresponding ground truth, and $\mathbf{W}$ denotes the set of parameters of the feature extractor while $\mathbf{w}^{(s)}$ refers to the parameters of all the layers at $s$th side-output. In addition, $P(\mathbf{G}(l) = 1| \mathbf{I}(l); \mathbf{W}, \mathbf{w}^{(s)})$ represents the probability of the activation value at location $l$ in the $s$th side output where $l$ is the spatial coordinate.

Then, the overall learning objective can be formulated as:
\begin{equation}
\min_{\mathbf{W}, \mathbf{w}}\sum^{S}_{s = 1}\mathcal{L}^{(s)}(\mathbf{I}, \mathbf{G}, \mathbf{W}, \mathbf{w}^{(s)}),
\end{equation}
where $S$ is the total side-output number, and $\mathbf{w}$ is the collections of  parameters of all layers at all side-ouputs, which is represented by 
\begin{equation}
\mathbf{w} = (\mathbf{w}^{(1)}, \mathbf{w}^{(2)}, \cdots, \mathbf{w}^{(S)}).
\end{equation}
In this work, $S$ is equal to 5.

%we denote the final output of $i$th side-output $\mc{R}_{i} \ (i= 1,2, \cdots, 5)$ as $P(\theta)^i$ where $\theta$ is the set of parameters of the whole network. Moreover, the ground-truth mask of input image is denoted as $G$. Then, the loss of the prediction of $i$th side-output is defined as
%\begin{equation}
%\mathcal{L}^i = E(\sigma(P(\theta)^i), G),
%\end{equation}
%where $\theta$ is sigmoid function. 

\section{Expriments}
In this section, we benchmark the proposed approach \textbf{DDS} and other 13 state-of-the-art SOD methods on the proposed 360$^\circ$ dataset \textbf{360-SOD}. 

\subsection{Experimental Settings.}

\textbf{1) Dataset.}
In the comparisons, we divide \textbf{360-SOD} into two subsets: 80\% for training (400 images) and 20\% for testing (100 images) by random shuffle algorithm.

\textbf{2) Evaluation metrics.}
We adopt mean absolute error (MAE), F-measure score (F$_{\beta}$) and weighted F-measure score (F$^w_{\beta}$) \cite{margolin2014evaluate} as our evaluation metrics. MAE reflects the average pixel-wise absolute difference between the estimated and ground-truth saliency maps. In computing F$_{\beta}$, we normalize the predicted saliency maps into the range of [0, 255] and binarize the saliency maps with a threshold sliding from 0 to 255 to compare the binary maps with ground-truth maps. At each threshold, Precision and Recall can be computed. F$_{\beta}$ is computed as:
\begin{equation}\label{eq:Fbeta}
F_{\beta} = \frac{(1 + \beta^2) \cdot Precision \cdot Recall}{\beta^2 \cdot Precision + Recall}.
\end{equation}
where we set $\beta^2 = 0.3$ to emphasize more on Precision than Recall as suggested in \cite{achanta2009frequency}. We report F$_{\beta}$ using an adaptive threshold for generating binary a saliency map and the threshold is computed as twice the mean of a saliency map. Besides, F$^w_{\beta}$ is computed to reflect the overall performance (refer to \cite{margolin2014evaluate} for details).

\textbf{3) Training and inference.}
We use standard stochastic gradient descent algorithm to train the whole network end-to-end with the cross-entropy losses between estimated saliency maps and ground-truth masks. In the optimization process, the parameter of feature extractor is initialized by the pre-trained ResNet-50 model \cite{he2016deep}, whose learning rate is set to $5 \times 10^{-9}$ with a weight decay of 0.0005 and momentum of 0.9. The learning rates of the rest layers in our network are set to 10 times larger. Besides, we employ the ``poly'' learning rate policy for all experiments similar to \cite{liu2015parsenet}. 

We train our network by utilizing the training set of \textbf{360-SOD}, which comprises of per-pixel ground-truth annotation for 400 images. The training images are resized to the resolution of $512 \times 256$ with the treatment of horizontal flipping. In our experiment, we cut the input image into $4 \times 4$ image blocks. The training process takes about 1.5 hours and converges after 50k iterations with mini-batch of size 1 on a single GTX 1080ti GPU. During testing, the proposed network removes all the losses, and each image is directly fed into the network to obtain its saliency map at the first side-output without any pre-processing. Due to the limitation of GPU memory and to improve the training efficiency, we downsample to a maximum side length of 512 pixels to conduct all experiments.

\subsection{Model Benchmarking}

\begin{figure*}[t]
\centering
\includegraphics[width=0.98\textwidth]{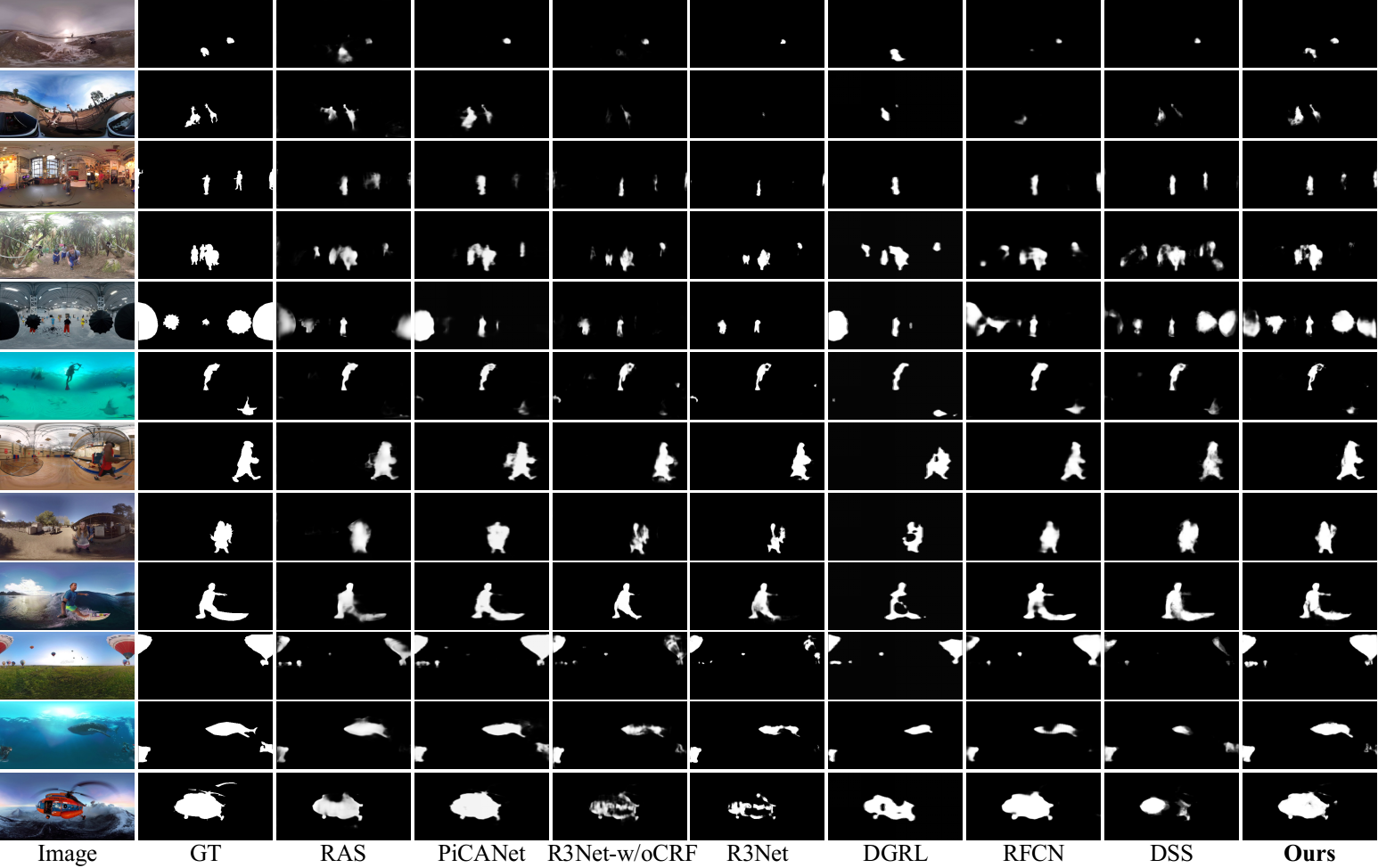}
\caption{Representative examples of the state-of-the-art algorithms after being fine-tuned on \textbf{360-SOD}. ``R3Net-wo'' means R3Net without dense CRF~\cite{krahenbuhl2011efficient}.}
\label{fig:results-after}
\end{figure*}

\begin{table}[t]
\centering
\caption{Performance of \textbf{DDS} and the state-of-the-art models after being fine-tuned on \textbf{360-SOD}. Note ``R3Net-w/ocrf'' means R3Net without dence CRF\cite{krahenbuhl2011efficient} and ``-'' means the training code is not available. The best three results are in {\color{red}{\textbf{\underline{red}}}}, {\color{green}{\textbf{\underline{green}}}} and {\color{blue}{\textbf{\underline{blue}}}}.}
\setlength{\tabcolsep}{5.4mm}{
\renewcommand\arraystretch{1.2}
\begin{tabular}{c | c c c}
\hline
 & MAE $\downarrow$
 & F$^w_\beta$ $\uparrow$
 & F$_\beta$ $\uparrow$
 \\
\hline
ELD\cite{lee2016deep} & - & - & - \\
UCF\cite{zhang2017learning} & 0.046 & 0.353 & 0.372 \\
NLDF\cite{luo2017non} & 0.042 & 0.402 & 0.424 \\
Amulet\cite{zhang2017amulet} & 0.031 & 0.526 & 0.583  \\
FSN\cite{chen2017look} & 0.030 & 0.529 & 0.609 \\
SRM\cite{wang2017stagewise} & 0.028 & 0.538 & 0.593 \\
C2SNet\cite{li2018contour} & - & - & -\\
RAS\cite{chen2018reverse} & 0.031 & 0.555 & 0.537 \\
PiCANet\cite{liu2018picanet} & {\color{blue}{\textbf{\underline{0.026}}}} & 0.578 & 0.589 \\
R3Net-w/oCRF\cite{deng2018r3net} & 0.029 & 0.546 & 0.599 \\
R3Net\cite{deng2018r3net} & 0.028 & 0.551 & {\color{red}{\textbf{\underline{0.677}}}} \\
DGRL\cite{wang2018detect} & 0.042 & 0.427 & {\color{blue}{\textbf{\underline{0.630}}}} \\
RFCN\cite{wang2018salient} & 0.027 & {\color{blue}{\textbf{\underline{0.585}}}} & 0.603 \\
DSS\cite{hou2019deeply} & {\color{green}{\textbf{\underline{0.025}}}} & {\color{green}{\textbf{\underline{0.591}}}} & 0.599 \\
\textbf{Ours}
& {\color{red}{\textbf{\underline{0.023}}}} & {\color{red}{\textbf{\underline{0.652}}}} & {\color{green}{\textbf{\underline{0.650}}}}  \\ 
\hline
\end{tabular}}
\label{tab:performance_after}
\end{table}

To show the challenges of \textbf{360-SOD}, we list the state-of-the-art model performance in Tab.~\ref{tab:performance_before} before fine-tuning them on \textbf{360-SOD}. These models include ELD~\cite{lee2016deep}, UCF~\cite{zhang2017learning}, NLDF~\cite{luo2017non}, Amulet~\cite{zhang2017amulet}, FSN\cite{chen2017look}, SRM~\cite{wang2017stagewise}, C2SNet~\cite{li2018contour}, RAS~\cite{chen2018reverse}, PiCANet~\cite{liu2018picanet}, R3Net~\cite{deng2018r3net}, DGRL~\cite{wang2018detect}, RFCN~\cite{wang2018salient} and DSS~\cite{hou2019deeply}.

On this dataset, DSS and RAS achieve the good performance, which maybe indicate the deep supervision is beneficial to the 360$^\circ$ image-based SOD. Moreover, R3Net and DGRL also obtain good results. R3Net is a method that combines high-level and low-level features, and dense conditional random field (dense CRF)~\cite{krahenbuhl2011efficient} are adopted for post-processing. DGRL is proposed to learn the local contextual information for each spatial position to refine boundaries. From these comparisons, we can believe the different-level features, different-scale contextual information and boundary refinement may be useful for saliency detection on omnidirectional image. 

Moreover, some representative examples are shown in Fig.~\ref{fig:results-before} for comprehensive analysis of these state-of-the-art on \textbf{360-SOD}. The figure shows best and worst examples of 8 state-of-the-art conventional SOD methods. We can observe that the large-scale salient objects in simple scene usually can be detected by all these approaches, while the small salient objects in complex scene will cause all approaches to fail.

Beyond the direct performance comparisons without fine-tuning, we fine-tune the proposed method and the state-of-the-art models on the training set of \textbf{360-SOD} dataset. The performance scores of the fine-tuned models on \textbf{360-SOD} are listed in Tab.~\ref{tab:performance_after}. Some representative results of \textbf{DDS} and other methods are shown in Fig.~\ref{fig:results-after}. Comparing Tab.~\ref{tab:performance_before} and Tab.~\ref{tab:performance_after}, we can observe that the performance of all the methods is all improved. In Tab.~\ref{tab:performance_after}, it is worth noting that F$^w_\beta$ of our method is significantly better compared with the second best results (0.652 against 0.591). Our method also consistently outperforms other models on MAE and F$_\beta$ except R3Net with dense CRF~\cite{krahenbuhl2011efficient}.
The dense CRF post-processes the predicted saliency maps to obtain sharper boundaries, which will improve performance, but also greatly increase the inference time. In Fig.~\ref{fig:results-after}, we can find the proposed method has better results comparing with other state-of-the-art methods on \textbf{360-SOD}.

From these experiments, we can believe that the proposed 360$^\circ$ image dataset \textbf{360-SOD} is a challenging dataset for omnidirectional scenes. Moreover, from the comparisons of the proposed baseline model and other state-of-the-art methods, we can know the proposed model has good performance and be useful for solving the problem of SOD.

\subsection{Performance Analysis of the baseline model}
\textbf{1) Generalization ability.}
\begin{table}[t]
\centering
\caption{Performance of \textbf{DDS} on randomly divided \textbf{360-SOD} for three more times.}
\setlength{\tabcolsep}{8mm}{
\renewcommand\arraystretch{1.2}
\begin{tabular}{c | c c c}
\hline
 & MAE $\downarrow$
 & F$^w_\beta$ $\uparrow$
 & F$_\beta$ $\uparrow$
 \\
\hline
1
& 0.022 & 0.665 & 0.651 \\
2
& 0.023 & 0.664 & 0.637 \\
3
& 0.023 & 0.654 & 0.644 \\
\hline
\end{tabular}}
\label{tab:generalization}
\end{table}
To validate the generalization ability of the proposed baseline model on \textbf{360-SOD}, we randomly re-divide the training and testing subsets on the 360$^\circ$ dataset and re-train \textbf{DDS}. This operation is conducted three times and the experimental results are shown in Tab.~\ref{tab:generalization}. From Tab.~\ref{tab:generalization}, we can believe \textbf{DDS} can stably achieve good performance on omnidirectional datasets.

\begin{table}[t]
\centering
\caption{Performance of \textbf{DDS} and the state-of-the-art models on \textbf{360-SOD-AT}. Note ``R3Net-w/ocrf'' means R3Net without dence CRF\cite{krahenbuhl2011efficient} and ``-'' means the training code is not available. The best three results are in {\color{red}{\textbf{\underline{red}}}}, {\color{green}{\textbf{\underline{green}}}} and {\color{blue}{\textbf{\underline{blue}}}}.}
\setlength{\tabcolsep}{5.4mm}{
\renewcommand\arraystretch{1.2}
\begin{tabular}{c | c c c}
\hline
 & MAE $\downarrow$
 & F$^w_\beta$ $\uparrow$
 & F$_\beta$ $\uparrow$
 \\
\hline
ELD\cite{lee2016deep} & - & - & - \\
UCF\cite{zhang2017learning} & 0.047 & 0.352 & 0.361 \\
NLDF\cite{luo2017non} & 0.041 & 0.408 & 0.419 \\
Amulet\cite{zhang2017amulet} & 0.031 & 0.508 & 0.581  \\
FSN\cite{chen2017look} & 0.030 & 0.564 & {\color{blue}{\textbf{\underline{0.620}}}} \\
SRM\cite{wang2017stagewise} & {\color{blue}{\textbf{\underline{0.027}}}} & 0.555 & 0.590 \\
C2SNet\cite{li2018contour} & - & - & -\\
RAS\cite{chen2018reverse} & 0.030 & 0.574 & 0.544 \\
PiCANet\cite{liu2018picanet} & 0.028 & {\color{blue}{\textbf{\underline{0.598}}}} & 0.582 \\
R3Net-w/oCRF\cite{deng2018r3net} & 0.030 & 0.557 & 0.598 \\
R3Net\cite{deng2018r3net} & 0.029 & 0.568 & {\color{red}{\textbf{\underline{0.675}}}} \\
DGRL\cite{wang2018detect} & 0.043 & 0.429 & {\color{blue}{\textbf{\underline{0.620}}}} \\
RFCN\cite{wang2018salient} & 0.028 & 0.589 & 0.603 \\
DSS\cite{hou2019deeply} & {\color{red}{\textbf{\underline{0.025}}}} & {\color{green}{\textbf{\underline{0.627}}}} & 0.605 \\
\textbf{Ours}
& {\color{red}{\textbf{\underline{0.025}}}} & {\color{red}{\textbf{\underline{0.656}}}} & {\color{green}{\textbf{\underline{0.641}}}}  \\ 
\hline
\end{tabular}}
\label{tab:generalization2}
\end{table}
In order to further verify the generalization ability of \textbf{DDS}, we collect 50 omnidirectional images based on Sports-360 dataset~\cite{hu2017deep} for annotating and construct a new testing dataset for additional testing (named as \textbf{360-SOD-AT}). It is worth noting that the distribution of the constructed dataset is relatively different from that of \textbf{360-SOD}. The dataset is constructed and annotated using the process described in Section III. \textbf{DDS} and other state-of-the-art models are evaluated on the new testing dataset and the results are listed in Tab.~\ref{tab:generalization2}. From Tab.~\ref{tab:generalization2}, we can see that proposed method perferms well and consistently outperforms other state-of-the-art methods, which validates the good generalization ability of the proposed method.

\textbf{2) Influence of various components.}
\begin{table}[t]
\centering
\caption{Performance of different settings about \textbf{DDS} on \textbf{360-SOD}. ``DA'' means distortion adaptation, ``MCI'' represents multi-scale context integration and ``DS'' is deep supervision. \#Parames: the number of parameters (Millions), \#FLOPS: floating point operations (Billions) and \#Time: average testing time (Millisecond).}
\setlength{\tabcolsep}{0.6mm}{
\renewcommand\arraystretch{1.2}
\begin{tabular}{c | c c c | c c c}
\hline
 & MAE $\downarrow$
 & F$^w_\beta$ $\uparrow$
 & F$_\beta$ $\uparrow$
 & \#Params
 & \#FLOPs
 & \#Time
 \\
\hline
FCN-dilation
& 0.025 & 0.611 & 0.632 & 27.2M & 59.7B & 52.0ms    \\
FCN-dilation + DA
& 0.025 & 0.623 & 0.637 & 27.2M & 59.8B & 53.5ms  \\
FCN-dilation + MCI
& 0.025 & 0.614 & 0.635 & 27.0M & 59.4B & 53.3ms  \\
FCN-dilation + DS
& \textbf{0.023} & 0.652 & 0.618 & 27.4M & 60.7B & 56.9ms  \\
FCN-dilation + DA + MCI
& 0.024 & 0.624 & 0.642 & 27.0M & 59.4B & 54.3ms  \\
FCN-dilation + DA + DS
& \textbf{0.023} & \textbf{0.658} & 0.633 & 27.4M & 60.8B & 57.7ms  \\
FCN-dilation + MCI+ DS
& 0.024 & 0.651 & 0.626  & 27.2M & 60.3B & 57.5ms  \\
\textbf{DDS}
& \textbf{0.023} & 0.652 & \textbf{0.650} & 27.2M & 60.4B & 59.0ms \\
\hline
\end{tabular}}
\label{tab:ablation}
\end{table}
To validate the effectiveness of different components of the proposed method, we conduct several experiments on \textbf{360-SOD}. Firstly, we construct a naive model with the only feature extractor and some conventional convolutional layers as described in Section \uppercase\expandafter{\romannumeral4} and shown in Fig.~\ref{fig:framework}, which is denoted as ``FCN-dilation''. Next, we construct three models that add different components to ``FCN-dilation'', including distortion-adaptive module (named as DA), multi-scale context integration module (named as MCI) and deep supervision (named as DS). Moreover, we combine these three components in pairs to obtain the other three models. These models are trained on the training set of \textbf{360-SOD} and the performance on the testing set of \textbf{360-SOD} is shown in Tab.~\ref{tab:ablation}. From Tab.~\ref{tab:ablation}, we can find deep supervision is useful for the improvement of F$^w_\beta$ by comparing settings with and without ``DS''. Moreover, it is easy to find that ``DA'' and ``MCI'' can stably promote the performance of \textbf{DDS}. 

In addition, we analyze the parameters, floating points operations (FLOPs) and average testing time of ablation models as listed in Tab.~\ref{tab:ablation}. From Tab.~\ref{tab:ablation}, we can observe that DA boosts the performance with slight additional parameters, FLOPs and time consumption. Meanwhile, MCI reduces the amount of parameters and FLOPs due to the reduction of channels (from 512 to 128), and improves the performance, but results in a little of time consumption. Also, DS obviously improves the performance while some additional parameters, FLOPs and time are used. 
From the above analysis, we can see that each component can improve performance, but extra parameters and computation are introduced. In fact, this performance improvement is not caused by simple parameters or FLOPs increase and an obvious example is that MCI improves its performance with parameters and FLOPs decrease. Therefore, it can be considered that the design of each component is effective and efficient. Finally, our model integrates the three proposed components, which has obvious performance improvement compared with the naive model ``FCN-dilation'', but only adds a small amount of FLOPs and testing time.

\section{Conclusion}
Due to the lack of omnidirectional datasets for salient object detection (SOD), the development of 360$^\circ$ SOD is restricted. To address this problem we propose the first public available 360$^\circ$ image-based SOD dataset. This dataset contains various indoor and outdoor scene from different perspectives as well as different scene complexities. Moreover, we explore the existing issues and find there exist three main problems that lead to the difficulty of SOD in 360$^\circ$ dataset, \ie distortion from projection, large-scale complex scene and small salient objects. Inspired by these issues, a baseline model is proposed on 360$^\circ$ SOD. This model organized in a progressive manner, utilizes distortion-adaptive and multi-scale context integration module to deal with existing problems. In addition, we provide a comprehensive benchmark of our approach and other 13 state-of-the-art conventional SOD algorithms on the proposed 360$^\circ$ SOD dataset, which shows the key challenges in omnidirectional scenes and validates the usefulness of the proposed dataset and baseline model. We believe that the dataset and the baseline model are helpful for the development of 360$^\circ$ SOD.

\section*{Acknowledgments}
This work was partially supported by grants from National Natural Science Foundation of China (61672072, 61922006, 61825101 and 61532003) and the Beijing Nova Program (Z181100006218063).

\bibliographystyle{IEEEtran}
\bibliography{Ref360-SOD_simple}
%\vspace{-2in}

\begin{IEEEbiography}[{\includegraphics[width=1in,height=1.25in,clip,keepaspectratio]{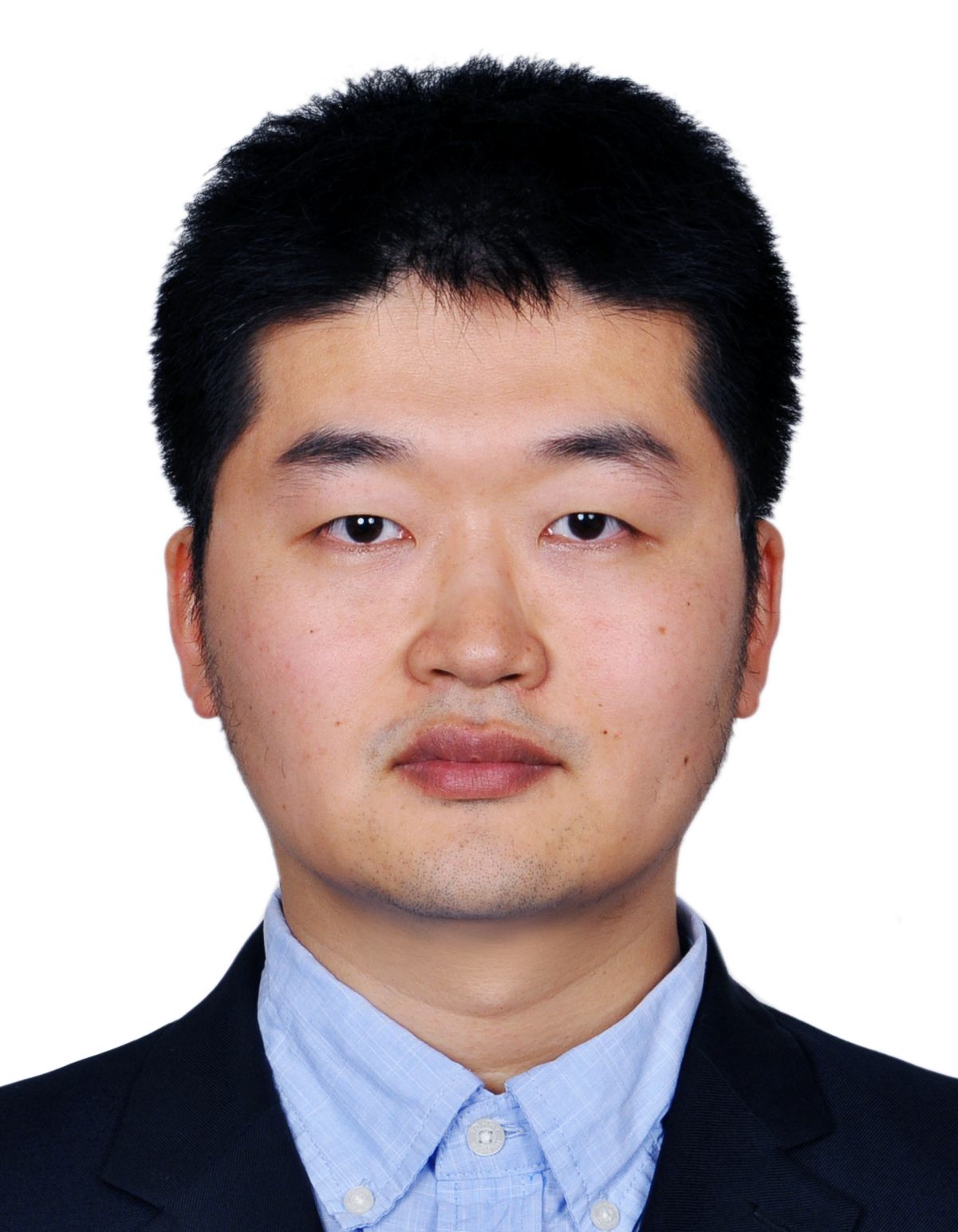}}]{Jia Li}
is currently an associate Professor with the School of Computer Science and Engineering, Beihang University, Beijing, China. He received the B.E. degree from Tsinghua University in Jul. 2005 and the Ph.D. degree from the Institute of Computing Technology, Chinese Academy of Sciences, in Jan. 2011. Before he joined Beihang University, he used to serve in Nanyang Technological University, Peking University and Shanda Innovations. His research interests include computer vision and multimedia big data, especially the cognitive vision towards evolvable algorithms and models. He is the author or coauthor of over 60 technical articles in refereed journals and conferences such as TPAMI, IJCV, TIP, ICCV and CVPR. More information can be found at http://cvteam.net.
\end{IEEEbiography}
%\vspace{-2in}

\begin{IEEEbiography}[{\includegraphics[width=1in,height=1.25in,clip,keepaspectratio]{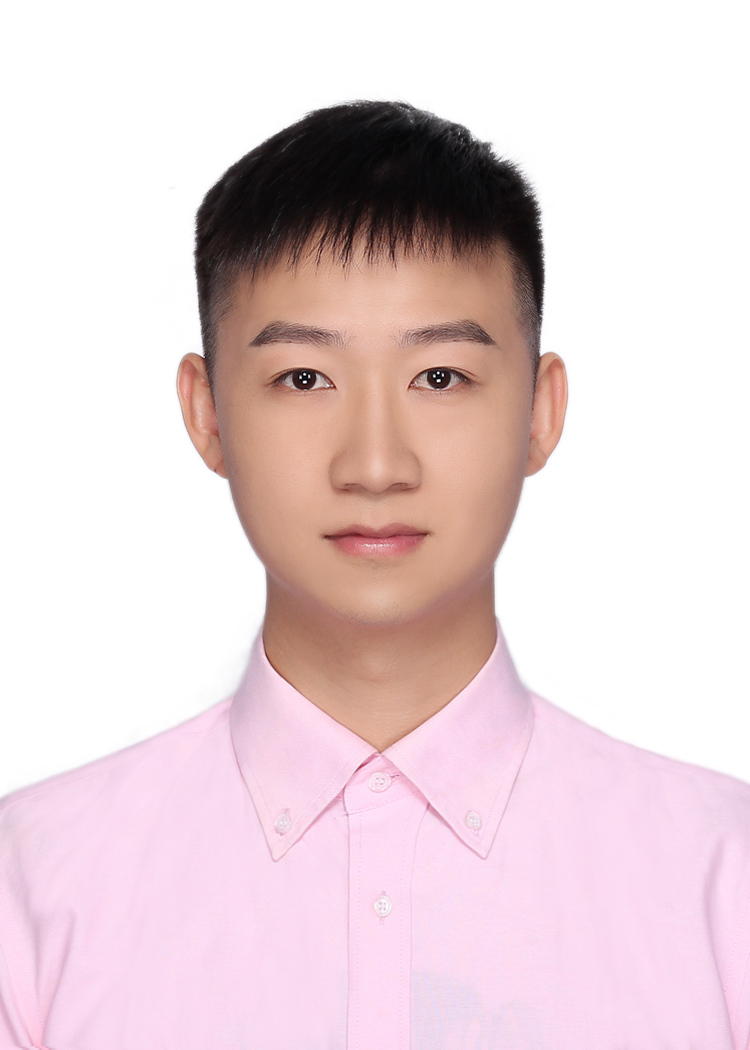}}]{Jinming Su}
is currently pursuing his master degree with the State Key Laboratory of Virtual Reality Technology and Systems, School of Computer Science and Engineering, Beihang University. His research interests include computer vision and deep learning.
\end{IEEEbiography}
%\vspace{-2in}

\begin{IEEEbiography}[{\includegraphics[width=1in,height=1.25in,clip,keepaspectratio]{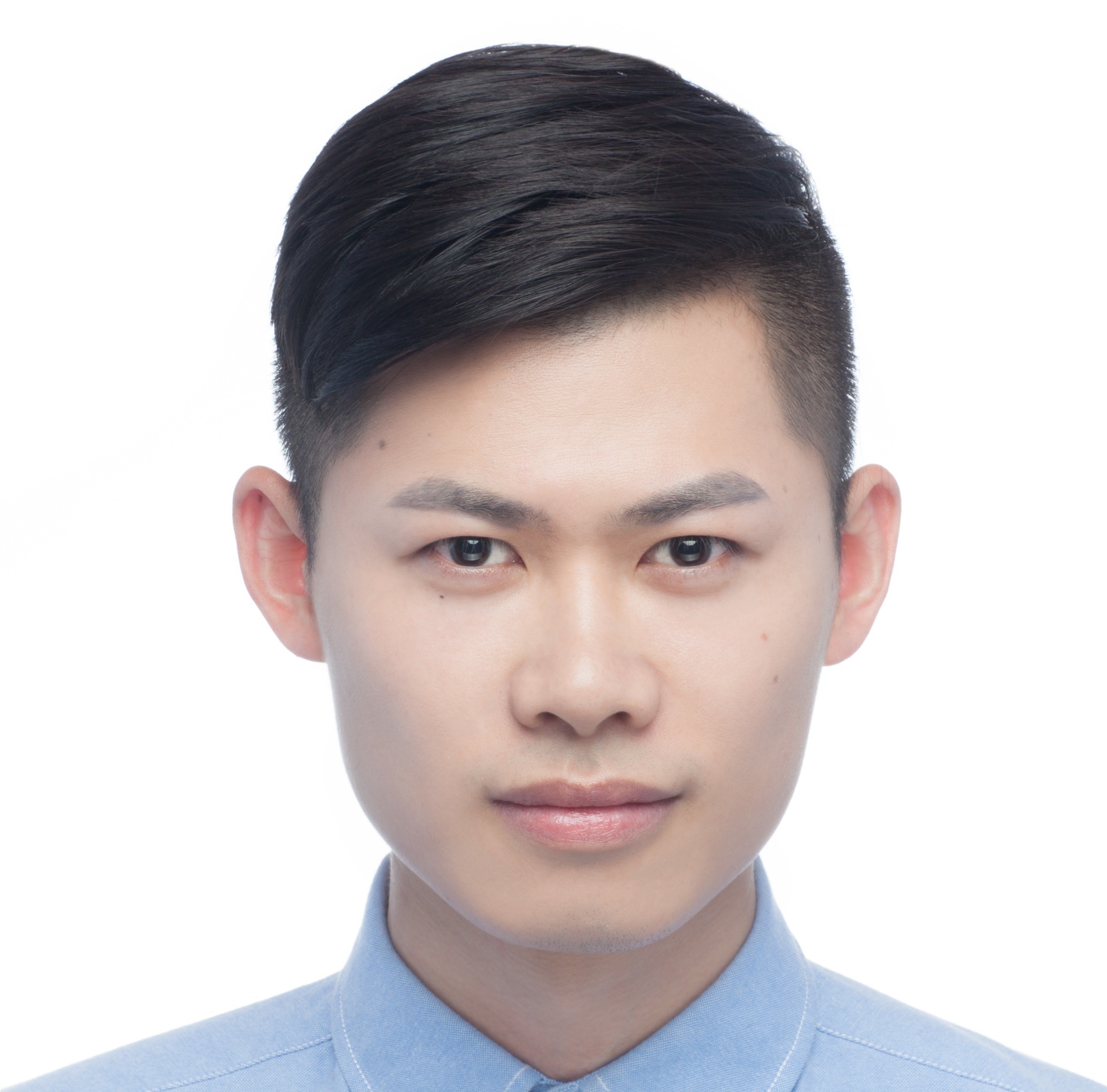}}]{Changqun Xia}
is currently an assistant Professor at Peng Cheng Laboratory, China. He received the Ph.D. degree from the State Key Laboratory of Virtual Reality Technology and Systems, School of Computer Science and Engineering, Beihang University, in Jul. 2019. His research interests include computer vision and image understanding.
\end{IEEEbiography}
%\vspace{-2in}

\begin{IEEEbiography}[{\includegraphics[width=1in,height=1.25in,clip,keepaspectratio]{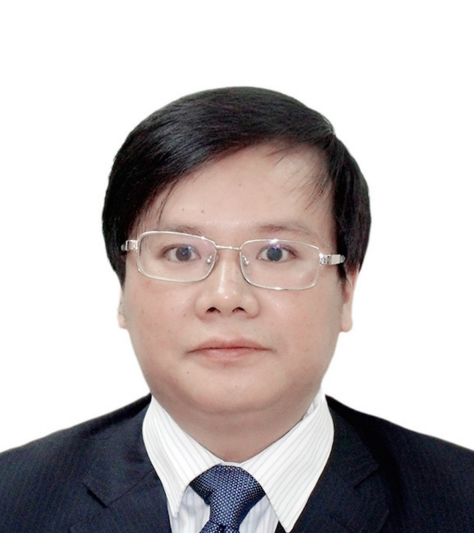}}]{Yonghong Tian}
is a full Professor with the School of Electronics Engineering and Computer Science, Peking University, Beijing, China. His research interests include machine learning, computer vision, and multimedia big data. He is the author or coauthor of over 140 technical articles in refereed journals and conferences and has owned more than 40 patents.
\end{IEEEbiography}
%\vspace{-2in}
\end{document}